\documentclass[letterpaper]{article} % DO NOT CHANGE THIS
\usepackage{aaai2027}  % DO NOT CHANGE THIS
\usepackage[hyphens]{url}  % DO NOT CHANGE THIS
\usepackage{graphicx} % DO NOT CHANGE THIS
\usepackage{natbib}  % DO NOT CHANGE THIS AND DO NOT ADD ANY OPTIONS TO IT
\usepackage{caption} % DO NOT CHANGE THIS AND DO NOT ADD ANY OPTIONS TO IT
\usepackage{algorithm}
\usepackage{algorithmic}

\usepackage{amssymb}
\usepackage{amsmath}

\usepackage{newfloat}
\usepackage{listings}
\DeclareCaptionStyle{ruled}{labelfont=normalfont,labelsep=colon,strut=off} % DO NOT CHANGE THIS
\floatstyle{ruled}
\newfloat{listing}{tb}{lst}{}
\floatname{listing}{Listing}

\usepackage{booktabs}

\nocopyright
\title{Drawing-Recode: Annotation Grounding for Parametric CAD Code Generation from Raster 2D CAD Drawings}
\author{
    Mingi Kim, Yongjun Kim, Hyungki Kim\corresponding
}
\affiliations{
    Department of Computer Science and Engineering\\
    Chungnam National University\\
    Daejeon, Republic of Korea\\
    mingi@o.cnu.ac.kr, kimit@g.cnu.ac.kr, hk.kim@cnu.ac.kr
}

\begin{document}

\maketitle

\begin{abstract}
Recovering Parametric CAD sequences from raster-format 2D Computer-Aided Design (CAD) drawings accumulated prior to digital transformation is important for part reproduction and manufacturing process automation. However, existing studies either process only vector drawings or are limited to specific domains, and fail to explicitly connect dimensional annotations to geometric information, limiting their use of dimensional information for 3D Parametric CAD sequences recovery. We propose Drawing-Recode, a framework that generates Parametric CAD sequences as CAD code from raster 2D CAD drawings. Drawing-Recode extracts geometric features via an image encoder and recognizes annotations through a separate text recognition module, then explicitly grounds annotations to geometric information using cross-attention and our proposed Annotation Grounding Loss (AGL). The resulting features are fed into a Large Language Model (LLM) to generate CAD code in the Structured Parametric CAD Code (SPCC) format. Experiments show that Drawing-Recode outperforms existing baselines and remains robust on scanned drawings resembling industrial conditions. We expect Drawing-Recode contributes to digitizing raster 2D CAD drawings in industrial settings and to part reproduction and manufacturing automation.
\end{abstract}


\section{Introduction}

Computer-Aided Design (CAD) is a core technology underlying modern manufacturing, widely used in product design, verification, and production \cite{cad1, cad2, cad-survey}. In particular, Parametric CAD sequences, which represent 3D geometric information as sequences of commands, have become the industry standard for design modification and model generation \cite{deepcad, fusion360, cad-parser, cadcl}. Meanwhile, industrial settings hold vast amounts of 2D CAD drawings accumulated prior to digital transformation, stored as raster-format scanned images or PDFs, and recovering these into 3D Parametric CAD sequences is an important task for part reproduction and manufacturing process automation \cite{plankassembly, cad2program}.

Efforts to recover 3D models from 2D drawings have long been studied using algorithmic approaches \cite{algorithm1, algorithm2, algorithm3, algorithm4}. However, these methods are designed to recover 3D models from drawings consisting only of shape outlines, without information such as dimensions or tolerances, making them difficult to apply to drawings containing dimensional information as found in industrial settings. To overcome this limitation, recent studies have leveraged deep learning to generate Parametric CAD sequences from 2D CAD drawings containing dimensional information \cite{cad2program, drawing2cad}.

\subsubsection{Understanding 2D CAD Drawings.}
A 2D CAD drawing consists of a geometry layer, representing the shape outline, and an annotation layer, specifying actual dimensions and tolerances \cite{cad2program}. The geometry layer captures the overall 3D shape structure, while the annotation layer directly provides numerical information---such as key dimensions, lengths, and angles---necessary for 3D shape recovery. Effectively leveraging both layers is thus essential for recovering Parametric CAD sequence from a 2D CAD drawing.

\begin{figure}[t]
\centering
\includegraphics[width=1.0\columnwidth]{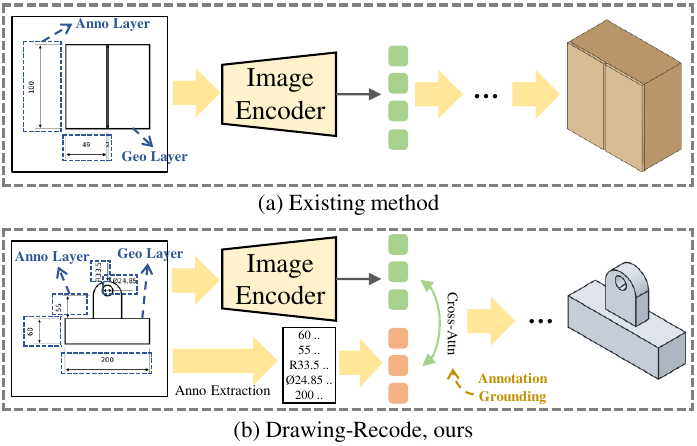}
\caption{(a) Existing methods, limited to a specific domain, jointly encode geometry and annotation layers. (b) Drawing-Recode extracts them independently and grounds annotations to the corresponding geometry.}
\label{fig1}
\end{figure}

To leverage raster drawings containing both layers, CAD2Program \cite{cad2program} proposed a framework that processes drawing images with a ViT-based image encoder \cite{vit} to generate Python-based CAD programs. However, CAD2Program processes drawings containing both the geometry layer and the annotation layer with a single image encoder, encoding the information of both layers into one feature, which limits its ability to learn the correspondence between the numerical information in the annotation layer and the geometry layer (Figure~\ref{fig1}(a)). It is also limited to a specific domain, cabinet furniture, making it difficult to apply to general mechanical part drawings.

In contrast, Drawing2CAD \cite{drawing2cad} successfully generated Parametric CAD sequences for mechanical part CAD models of various shapes based on the DeepCAD \cite{deepcad} dataset. However, it processes only vector drawings in Scalable Vector Graphics (SVG) format as input, making it inapplicable to CAD drawings stored in raster format.

To overcome these limitations, we propose Drawing-Recode. Drawing-Recode is a framework that takes raster-format 2D CAD drawings as input and generates a Parametric CAD sequences in the form of CAD code, addressing both limitations mentioned above. First, we construct a 2D CAD drawing dataset covering mechanical parts of diverse shapes based on the large-scale CAD dataset DeepCAD, enabling general CAD code generation not limited to a specific domain. Second, unlike the existing approach of processing both layers with a single image encoder, we design our framework to independently extract the geometry layer and the annotation layer and explicitly ground them, enabling explicit learning of the correspondence between the two layers (Figure~\ref{fig1}(b)).

Specifically, Drawing-Recode adopts a Large Language Model (LLM), widely used in recent CAD reverse engineering research, as its backbone to leverage pretrained coding ability, generating CAD code in the Structured Parametric CAD Code (SPCC) format, which enables stable generation of syntactically valid code \cite{cad-llama, recad, brepcoder}. It extracts geometry layer features via a CLIP ViT-based image encoder \cite{clip, vit}, while recognizing and embedding the annotation layer through YOLO \cite{yolo, yolov10} and SVTR \cite{svtr, svtrv2}, enabling independent extraction of features from the two layers. The two features are then combined through a cross-attention module, with our proposed Annotation Grounding Loss (AGL) training annotation layer features to explicitly align with their corresponding regions in the geometry layer. This allows the model to identify which part a given dimension corresponds to, enabling more precise 3D shape recovery. The resulting grounded features are fed into the LLM to generate Parametric CAD code. Our contributions are summarized as follows:
\begin{itemize}
    \item We construct a 2D CAD drawing dataset covering mechanical parts of diverse shapes based on the large-scale CAD dataset DeepCAD, enabling general CAD code generation not limited to a specific domain.
    \item We propose Drawing-Recode, a framework that independently extracts the geometry layer and the annotation layer, allowing the information of both layers to be represented according to their respective features.
    \item We design an explicit grounding mechanism consisting of a cross-attention module and our proposed Annotation Grounding Loss (AGL), training annotation features to be grounded to geometry features, enabling the generation of CAD code with precise 3D shape.
\end{itemize}

\section{Related Work}
\subsubsection{Drawing to CAD Generation.}
Research on generating CAD models from 2D drawings can be divided into algorithm-based methods and deep learning-based methods. Early studies leveraged mathematical frameworks and geometric reasoning to recover 3D models from three-view orthographic projections (Front, Right, Top) \cite{algorithm1, algorithm2, algorithm3, algorithm4}, but were limited to simple geometric shapes such as planes, cylinders, and spheres. Photo2CAD \cite{photo2cad} generated 3D models from 2D orthographic projections using an OpenCV-based algorithmic approach, but showed limited applicability to complex geometries. With the advancement of deep learning, learning-based methods subsequently emerged. Free2CAD \cite{free2cad} proposed a framework that generates CAD sequences from freehand sketches. PlankAssembly \cite{plankassembly} generated cuboid-based models from three-view vector drawings, but was limited to the furniture domain composed of planks. Drawing2CAD \cite{drawing2cad} proposed a seq2seq framework that generates Parametric CAD sequences from SVG-format vector drawings, and CAD2Program \cite{cad2program} processed raster drawing images with a ViT-based encoder \cite{vit, internvl} to generate Python-based CAD programs for cabinet models. However, these studies are limited to a specific SVG format or specific domains such as planks and cabinets, and fail to explicitly connect dimensional annotations in the drawing to geometric information, making it difficult to directly leverage dimensional information for 3D shape recovery.

\subsubsection{LLM-based CAD Code Generation.} 
Recently, active research has explored leveraging the code generation capabilities of Large Language Models (LLMs) for CAD generation. CAD-Recode \cite{cad-recode} combined an LLM with a point cloud projector to propose an LLM-based CAD reverse engineering model that converts point clouds into CadQuery-based Python code. cadrille \cite{cadrille} extended this by proposing a multimodal framework that jointly processes point clouds, images, and text, improving performance through reinforcement learning. For text-based input, Text-to-CadQuery \cite{text-to-cadquery} and CAD-Coder \cite{cad-coder} proposed methods that leverage LLMs to generate CadQuery-based code from natural language. Meanwhile, CAD-Llama \cite{cad-llama} proposed a custom Python-like format called Structured Parametric CAD Code (SPCC) to train a LLaMA-based model \cite{llama, llama3}, and ReCAD \cite{recad} adopted the SPCC format to generate CAD code conditioned on text and images, improving performance through reinforcement learning. BrepCoder \cite{brepcoder} is a framework that takes B-rep as input to generate CAD code in the SPCC format, proposing a pipeline that learns alignment between B-rep and code. While these studies leveraging the Python coding ability of LLMs for CAD generation have progressively advanced, none have yet been applied to research using 2D CAD drawings as input.

\section{Method}
We first introduce the overall pipeline of Drawing-Recode. Second, we describe the Drawing Encoder, which grounds annotations to geometry. Third, we describe the training loss. Finally, we describe dataset construction.

\begin{figure}[t]
\centering
\includegraphics[width=1.0\columnwidth]{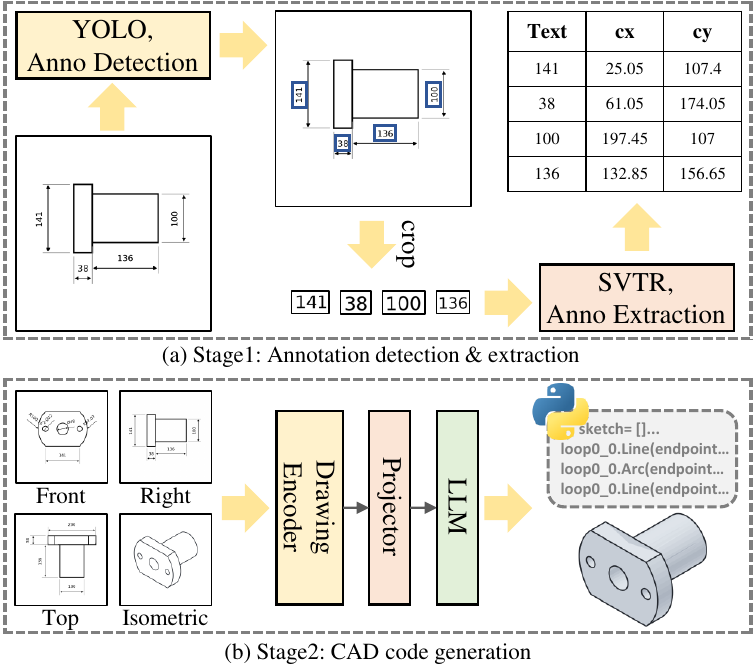}
\caption{Overall pipeline of Drawing-Recode. (a) Stage 1 detects and extracts annotation text via YOLO and SVTR. (b) Stage 2 encodes the four-view drawings with the Drawing Encoder and generates CAD code via the LLM.}
\label{fig2}
\end{figure}

\subsection{Overall Architecture}
The overall pipeline of Drawing-Recode consists of two stages, as shown in Figure~\ref{fig2}.

\subsubsection{Stage 1: Annotation Detection \& Extraction.} 
The first stage extracts the annotation layer from the input drawing. A YOLO-based detector first detects each annotation text region on the drawing as a bounding box \cite{yolo, yolov10}. The detected regions are cropped from the drawing into individual text images, which are then fed into an SVTR-based text recognition model to extract the numerical value of each annotation \cite{svtr, svtrv2}. This process yields an annotation table consisting of the recognized text and the center coordinates of the corresponding bounding box, and this module is pretrained prior to Stage 2.

\subsubsection{Stage 2: CAD Code Generation.} 
The second stage takes drawings from the Front, Right, Top, and Isometric views as input and generates CAD code. Each view is encoded through the Drawing Encoder into features in which the annotation layer is grounded to the geometry layer, and the detailed structure of the Drawing Encoder is described in the following section. Since the output of the Drawing Encoder differs in dimension from the input embedding space of the LLM, it is transformed into a dimension processable by the LLM through a Projector consisting of a single Linear layer. The transformed features are fed into the LLM to generate CAD code. The Drawing Encoder, Projector, and LLM are jointly trained in an end-to-end manner.

\subsection{Drawing Encoder}
\begin{figure*}[t]
\centering
\includegraphics[width=1.0\textwidth]{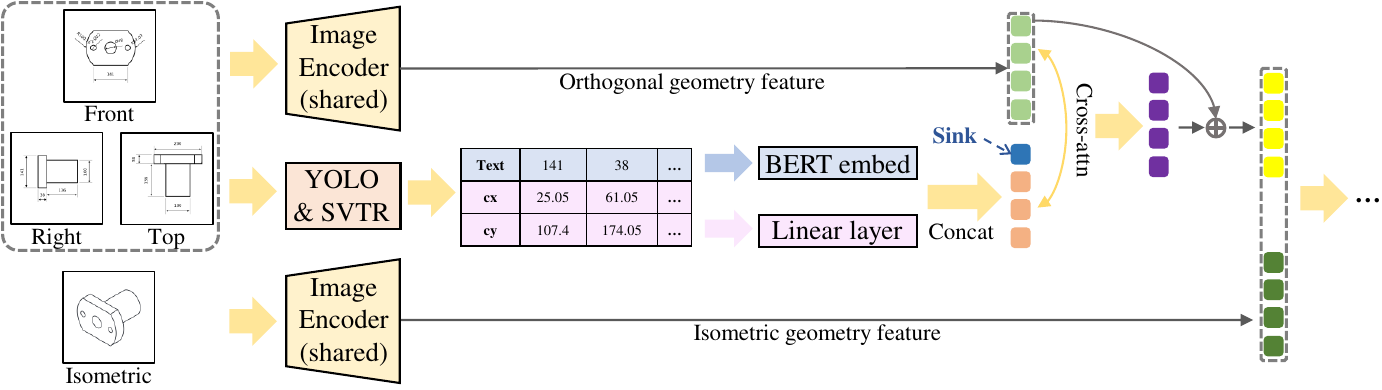}
\caption{Architecture of the Drawing Encoder. Geometry features from a shared image encoder are grounded via cross-attention to annotation features, which are embedded from the YOLO/SVTR annotation table together with a sink token, and then concatenated with the Isometric geometry feature.}
\label{fig3}
\end{figure*}

The Drawing Encoder is illustrated in Figure~\ref{fig3}.

\subsubsection{Geometry Feature Extraction.} 
Let $\mathcal{V}=\{F, R, T, Iso\}$  denote the set of Front, Right, Top, and Isometric views, respectively. The four view drawings $\{I_v\}_{v\in\mathcal{V}}$ at a resolution of $224\times224$ are fed into an image encoder with shared weights, where each view is encoded into $P=256$ patch features, which serve as the geometry features $G_v\in\mathbb{R}^{P\times1024}$ in this work. Here, 1024 is the hidden dimension of the encoder.

\subsubsection{Annotation Embedding.} 
For each orthogonal view $v_o\in\mathcal{V}_o$ $(\mathcal{V}_o=\{F, R, T\}\subset\mathcal{V})$, an annotation table $\{(t_i, c_{x, i}, c_{y, i})\}^{L_{v_o}}_{i=1}$ is extracted through the YOLO and SVTR models trained in Stage 1. Here, $t_i$ is the recognized text, $(c_{x, i}, c_{y, i})$ are the center coordinates of the corresponding text bounding box, and $L_{v_o}$ is the number of annotations in view $v_o$. The text $t_i$ is embedded into 768 dimensions via mean-pooled BERT token embeddings \cite{bert}, and the coordinates $(c_{x, i}, c_{y, i})$ are projected into 256 dimensions through a Linear layer. The two vectors are then concatenated to obtain a 1024-dimensional annotation feature $a_i$. The resulting $a_i$ is subsequently grounded, via cross-attention, to the patches of the geometry feature corresponding to each annotation.

\subsubsection{Sink Token.}
Not every patch in an orthogonal view corresponds to an annotation; there may be blank patches unrelated to any annotation. To account for such patches, we add a learnable sink token $s\in\mathbb{R}^{1024}$ to the annotation feature set. The final annotation feature set is thus $A_{v_o}=[s, a_1, \dots, a_{L_{v_o}}]\in\mathbb{R}^{(1+L_{v_o})\times 1024}$.

\subsubsection{Cross-attention Grounding.} 
For each orthogonal view $v_o$, we perform multi-head cross-attention with the geometry feature $G_{v_o}$ as Query and $A_{v_o}$ as Key and Value, as follows:
\begin{equation}
    C_{v_o}=\text{MultiHeadAttn}(G_{v_o}, A_{v_o}, A_{v_o}).
\end{equation}
Through this, each patch attends to its relevant annotations to retrieve information, while blank patches allocate their attention to the sink token, preventing specific annotation information from being injected into blank patches. The final orthogonal view feature is computed through a residual connection as $\hat{G}_{v_o}=G_{v_o}+C_{v_o}$.

The Front, Right, and Top views each undergo the above process independently to extract their features. The Isometric view is used as is, with $\hat{G}_{Iso}=G_{Iso}$, since it carries no dimensional annotations and instead provides complementary 3D shape context. Finally, the features of the four views are concatenated into $\hat{G}=\{\hat{G}_F;\hat{G}_R;\hat{G}_T;\hat{G}_{Iso}\}$, which serves as the output of the Drawing Encoder.

\subsection{Loss Function} 
The LLM is trained to autoregressively generate the target CAD code $y=(y_1, \dots,y_T)$ given the drawing features $\hat{G}$ as input. $\mathcal{L}_{LM}$ is defined as
\begin{equation}
    \mathcal{L}_{LM}=-\sum^T_{t=1}\log P(y_t\mid y_{<t}, \hat{G}).
\end{equation}
While $\mathcal{L}_{LM}$ supervises code generation, the grounding between the two layers is supervised directly on the cross-attention weights. The cross-attention weight $w_{v_o,p}\in\mathbb{R}^{1+L_{v_o}}$ for patch $p$ of orthogonal view $v_o$ is a probability distribution over $1+L_{v_o}$ entries including the sink token. The patch label $l_{v_o, p}$ is a one-hot vector denoting the annotation to which each patch corresponds. The Annotation Grounding Loss (AGL) is defined as the negative log-likelihood of $w_{v_o, p}$ with $l_{v_o, p}$ as the ground-truth label, averaged over the $P$ patches of each view and then averaged again over the orthogonal views.
\begin{equation}
    \mathcal{L}_{AGL}=\frac{1}{|\mathcal{V}_o|}\sum_{v_o\in\mathcal{V}_o}\left(-\frac{1}{P}\sum^P_{p=1}l_{v_o, p}\log w_{v_o, p}\right).
\end{equation}
The final loss is defined as a linear combination of the two terms,
\begin{equation}
    \mathcal{L}=\mathcal{L}_{LM} + \alpha \cdot \mathcal{L}_{AGL},
\end{equation}
where $\alpha$ is a weighting hyperparameter for the AGL.

\subsection{Dataset Generation}
Structured Parametric CAD Code (SPCC) is Python-like CAD code that generates parametric 3D solids, and has been widely adopted in recent studies on CAD sequence generation with LLMs \cite{cad-llama, recad, brepcoder}. In this work, we construct our dataset based on the DeepCAD dataset converted into the SPCC format. We execute the CAD code of each model to obtain a 3D solid, and then orthographically project it along the Front, Right, Top, and Isometric directions to generate SVG vector drawings. In this process, we do not normalize the drawings to a fixed canvas ratio but instead reflect the world coordinates of the model directly in the SVG, thereby fully preserving the dimensional information. The resulting SVG vector drawings are rendered with matplotlib into raster images at a resolution of $224\times224$ containing both the geometry layer and the annotation layer.

We use CLIP ViT-L/14 \cite{clip, vit} for geometry layer feature extraction, which divides the input image into a $16\times16$ grid of patches. To ground the annotation layer to the geometry layer, we additionally construct patch labels specifying which annotation value each patch corresponds to. For example, a patch corresponding to a line segment of length 200 is assigned a patch label of 200. Separately, we also extract bounding box information indicating the location of each annotation text on the drawing, for use in extracting the annotation layer. Further details on dataset construction, including SPCC conversion and drawing generation, are described in the Appendix.

\section{Experiments}
\subsection{Experimental Setups}
\subsubsection{Baselines.}
We compare Drawing-Recode against Drawing2CAD-vector, DeepCAD-raster, DeepCAD-vector, and CAD2Program \cite{deepcad, drawing2cad, cad2program}. Drawing2CAD-vector generates Parametric CAD sequences from SVG vector drawings. DeepCAD-raster and DeepCAD-vector pair the DeepCAD decoder with a raster image encoder and the Drawing2CAD SVG encoder, respectively. CAD2Program generates CAD programs from raster drawings with a vision-language model. Following its original setting, we adopt Mini-InternVL-1.5-2B \cite{internvl, mini-internvl, internvl1.5} as the backbone with an input resolution of $448 \times 448$, and train it on our dataset.

\subsubsection{Implementation Details.}
We use Qwen2.5-0.5B \cite{qwen, qwen2, qwen2.5} as the LLM backbone and adopt CLIP ViT-L/14 \cite{clip, vit} as the image encoder for geometry layer feature extraction. For annotation layer extraction with YOLO \cite{yolo, yolov10} and SVTR \cite{svtr, svtrv2}, we use YOLO26s-OBB and SVTR-small, respectively, initialized from pretrained weights, with SVTR trained on the crops produced by the trained detector. BERT-base \cite{bert} is used for annotation text embedding, and the Projector consists of a single Linear layer. Drawing-Recode has 0.85B parameters in total, including the YOLO and SVTR modules. On the test set, YOLO and SVTR achieve 99.55\% detection F1 and 98.54\% word accuracy. Training details for our model and the baselines are in the Appendix.

\subsubsection{Metrics.}
To evaluate the agreement between the generated CAD code and the ground truth, we use $\text{ACC}_{cmd}$ and $\text{ACC}_{param}$. $\text{ACC}_{cmd}$ measures the ratio of correctly predicted commands, while $\text{ACC}_{param}$ computes the ratio of parameters whose absolute error is within a tolerance $\delta$, only for correctly predicted commands.
\begin{align}
    \text{ACC}_{cmd}&=\frac{1}{N_c}\sum^{N_c}_{i=1}\mathbb{I}(\hat{c}_i=c_i),\\
    \text{ACC}_{param}&=\frac{1}{K}\sum^{N_c}_{i=1}\sum^{N_p}_{j=1}\mathbb{I}(|\hat{p}_{i, j}-p_{i, j}|<\delta) \cdot \mathbb{I}(\hat{c}_i=c_i),
\end{align}
where $N_c$ and $K$ are the total number of commands and parameters, $\mathbb{I}(\cdot)$ is the indicator function, and $c_i$ ($p_{i,j}$) and $\hat{c}_i$ ($\hat{p}_{i,j}$) denote the ground truth and prediction of the $i$-th command (its $j$-th parameter). To evaluate 3D shape quality, following Drawing2CAD \cite{drawing2cad}, we sample 2,000 points from the surface of the generated CAD model and compute the Mean Chamfer Distance (MCD). We also report the Invalid Ratio (IR) for CAD sequences failing 3D shape reconstruction.

\begin{table}[t]
  \centering
  \setlength{\tabcolsep}{0.6mm}
  \small
  \begin{tabular}{ccccc}
    \toprule
    Method & ACC$_{cmd} \uparrow$ & ACC$_{param} \uparrow$ & MCD $\downarrow$ & IR $\downarrow$ \\
    \midrule
    DeepCAD-raster   & 77.69 & 70.49 & 18.16 & 29.79 \\
    DeepCAD-vector  & 81.51 & 75.14 & 11.37  & 23.40 \\
    Drawing2CAD-vector     & 82.43 & 76.09 & 10.88 & 20.31 \\
    CAD2Program & 84.80 & 79.57 & 10.13 & 2.59 \\
    \textbf{Drawing-Recode} & \textbf{87.01} & \textbf{82.53} & \textbf{7.86} & \textbf{0.97} \\
    \bottomrule
  \end{tabular}
  \caption{Quantitative comparison between Drawing-Recode and the baselines. Mean Chamfer Distance (MCD) is scaled by $10^{2}$. The best results are highlighted in \textbf{bold}.}
  \label{tab1}
\end{table}

\subsection{Drawing to CAD Generation}
Table~\ref{tab1} shows the quantitative comparison between Drawing-Recode and the baselines. Drawing-Recode outperforms all baselines across all metrics. While vector drawings provide the coordinates of line segments directly as numerical values and raster drawings provide only pixel information, Drawing-Recode still outperforms Drawing2CAD-vector, improving MCD by 27.8\%. This indicates that independently extracting the geometry layer and the annotation layer and explicitly grounding them compensates for the limited information available in raster drawings. IR also decreases substantially from 20.31\% to 0.97\%, since SPCC is a Python-like format and generating it leverages the Python coding ability acquired by the LLM during pretraining, yielding syntactically valid and executable code.

\begin{figure*}[t]
\centering
\includegraphics[width=1.0\textwidth]{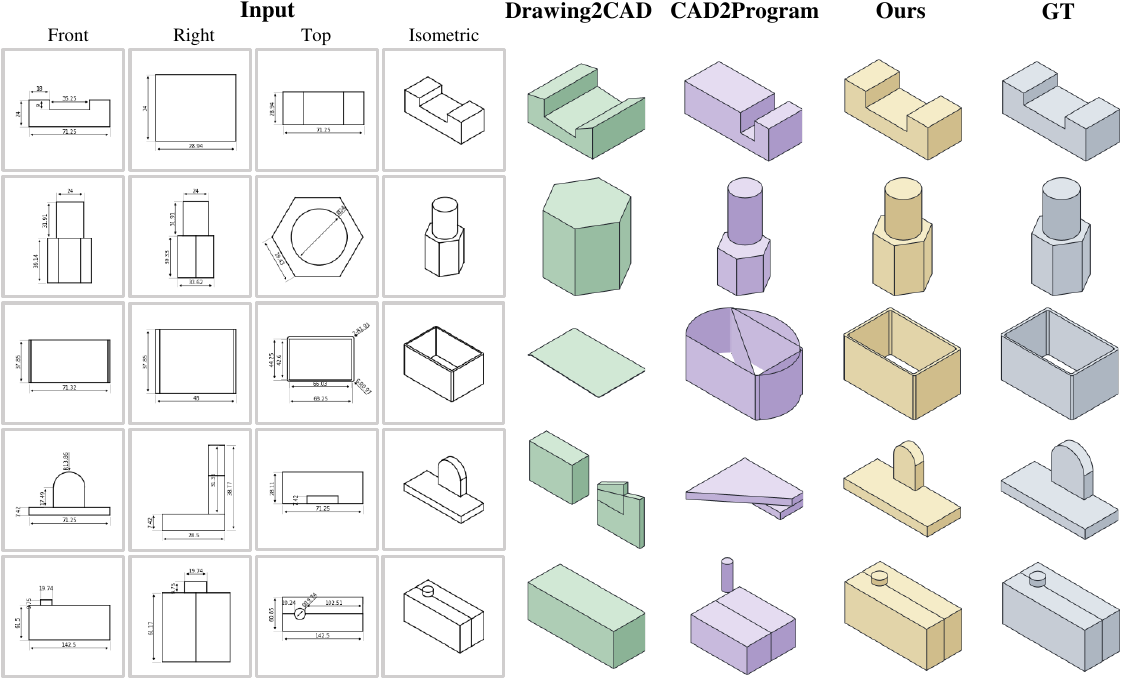}
\caption{Qualitative results. Given Front, Right, Top, and Isometric view drawings as input, we compare the 3D shapes reconstructed by Drawing2CAD-vector, CAD2Program, and Drawing-Recode with the ground truth.}
\label{fig4}
\end{figure*}

CAD2Program, which also generates CAD code with an LLM and thus benefits from the same pretrained Python coding ability, likewise achieves a low IR of 2.59\%. However, CAD2Program still falls behind Drawing-Recode across all metrics despite using a larger backbone (2B vs. 0.85B) and a higher input resolution ($448 \times 448$ vs. $224 \times 224$). This suggests that the improvement of Drawing-Recode stems from explicitly grounding annotations to geometry rather than from model capacity or input resolution. Qualitative results are shown in Figure~\ref{fig4}.

\begin{table}[t]
  \centering
  \setlength{\tabcolsep}{1.0mm}
  \small
  \begin{tabular}{ccccc}
    \toprule
    Method & ACC$_{cmd} \uparrow$ & ACC$_{param} \uparrow$ & MCD $\downarrow$ & IR $\downarrow$ \\
    \midrule
    w/o $(c_x,c_y)$   & 86.55 & 81.80 & 8.12 & 1.17 \\
    w/o Isometric  & 86.73 & \textbf{82.58} & \textbf{7.84} & 2.20 \\
    w/o Weight Sharing & 86.86 & 82.39 & 8.00 & 1.29 \\
    w Text Prompt & 71.98 & 69.73 & 9.66 & 2.89 \\
    w $\alpha=0.01$ & 86.73 & 82.13 & 8.14 & 1.33 \\
    w $\alpha=1.0$ & 86.55 & 81.38 & 8.28 & 1.32 \\
    \midrule
    Vanilla    & 80.51 & 71.37 & 9.20 & 3.48 \\
    + Cross-Attn & 86.29 & 81.42 & 8.23 & 1.58 \\
    + Sink & 86.23 & 81.41 & 8.01 & 1.59 \\
    \midrule
    \textbf{Ours (+AGL)}   & \textbf{87.01} & 82.53 & 7.86 & \textbf{0.97} \\
    \bottomrule
  \end{tabular}
  \caption{Ablation results. The upper block ablates individual design components, and the lower block adds each component sequentially to the Vanilla configuration.}
  \label{tab2}
\end{table}

\subsection{Ablation Studies}
\subsubsection{Design Choices.}
The upper block of Table~\ref{tab2} shows ablation results on the design components of Drawing-Recode. First, removing the bounding box center coordinates $(c_x, c_y)$ and performing cross-attention with only the numerical annotation information degrades performance. This indicates that the coordinates contribute to accurate grounding and thereby to performance. Second, excluding the Isometric view keeps ACC and MCD comparable to Ours while IR differs by 1.23\%. This suggests that the Front, Right, and Top views alone contain all the dimensional annotations and thus suffice for grounding, keeping ACC and MCD at a similar level. In contrast, the Isometric view contributes to generating syntactically valid and geometrically plausible code, so excluding it increases IR. Third, using separate CLIP ViT-L/14 encoders for the Front, Right, and Top views and for the Isometric view rather degrades performance. Pretrained at scale to represent both shape-only images and images containing text and numerical information, its 0.3B parameters appear sufficient to cover both within a single encoder. Fourth, providing annotations as a text prompt, instead of grounding them to geometry via cross-attention and the AGL, degrades performance. Given the same dimensions and coordinates, the model appears to rely on the numbers in the prompt rather than learning which geometry they correspond to. This shows that explicit grounding contributes more to performance than merely supplying the annotation information. Fifth, when the weight $\alpha$ of $\mathcal{L}_{AGL}$ is set 10 times smaller or larger than 0.1, performance degrades in both cases: with $\alpha=0.01$, $\mathcal{L}_{AGL}$ does not converge sufficiently, making accurate annotation grounding difficult, while with $\alpha=1.0$, the convergence of $\mathcal{L}_{AGL}$ improves but it interferes with the training of $\mathcal{L}_{LM}$, degrading the quality of the generated CAD code. The training curves of $\mathcal{L}_{AGL}$ and $\mathcal{L}_{LM}$ for different values of $\alpha$ are provided in the Appendix.

\begin{figure}[t]
\centering
\includegraphics[width=1.0\columnwidth]{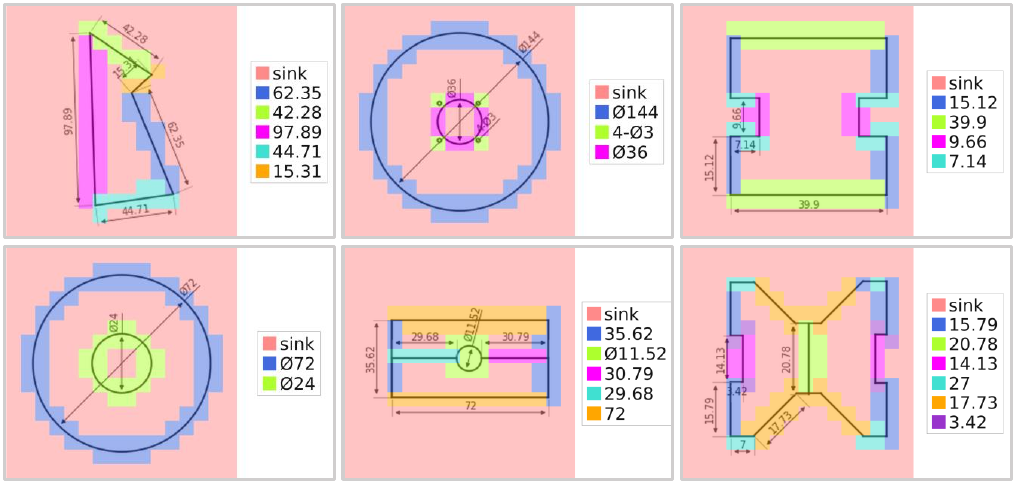}
\caption{Visualization of annotation grounding. Each patch is colored by the annotation it attends to most in the cross-attention module, and patches without a corresponding annotation are assigned to the sink token.}
\label{fig5}
\end{figure}

\subsubsection{Component-wise Addition.}
The lower block of Table~\ref{tab2} shows the results of sequentially adding each component. Starting from Vanilla, which consists only of the Image Encoder, Projector, and LLM, performance improves progressively across all metrics as the cross-attention module, the sink token, and the AGL are added in turn, with Ours (+AGL), which includes all components, achieving the best performance. In particular, the largest improvement is observed when the cross-attention module alone is added to Vanilla. This shows that processing the geometry layer and the annotation layer separately contributes substantially to the performance improvement. Qualitative results of annotations being grounded to each patch in the cross-attention module are shown in Figure~\ref{fig5}. Additional experiments using a different LLM backbone are described in the Appendix.

\begin{table}[t]
  \centering
  \setlength{\tabcolsep}{1.0mm}
  \small
  \begin{tabular}{ccccc}
    \toprule
    Method & ACC$_{cmd} \uparrow$ & ACC$_{param} \uparrow$ & MCD $\downarrow$ & IR $\downarrow$ \\
    \midrule
    w CadQuery & - & - & 0.84 & 1.57 \\
    \textbf{Ours}   & 87.01 & 82.53 & \textbf{0.76} & \textbf{0.97} \\
    \bottomrule
  \end{tabular}
  \caption{Comparison between the SPCC and CadQuery output formats. MCD is computed after center alignment, following the convention of CadQuery-based work.}
  \label{tab3}
\end{table}

\subsubsection{SPCC vs. CadQuery.}
Recent studies on LLM-based CAD generation widely adopt CadQuery as an output format in addition to SPCC. To examine whether the proposed grounding design depends on a specific code format, we construct a 2D CAD drawing dataset in the same manner from a publicly available CadQuery-format dataset and train our model on it. We evaluate with MCD and IR. Following the convention of prior work adopting the CadQuery format \cite{cadrille, cad-coder, cadcodeverify, futurecad}, we compute CD after aligning the centers of the generated and ground-truth models to the origin. This alignment yields MCD values in Table~\ref{tab3} on a different scale from those in Tables~\ref{tab1} and \ref{tab2}.

As shown in Table~\ref{tab3}, MCD remains at a comparable level with the CadQuery format. IR, on the other hand, is slightly higher at 1.57\% with CadQuery than the 0.97\% with SPCC. Analyzing the failure cases, all failures with SPCC stem from geometric errors such as unclosed loops, whereas 16.8\% of the failures with CadQuery are execution-stage errors unrelated to geometry, including RuntimeError, SyntaxError, and timeout. This appears to be because SPCC parses a small set of primitives directly with a custom parser, while CadQuery is executed through a library layer built on top of a CAD kernel, which introduces more points at which failures can occur. The comparable performance suggests that the code representation is an implementation choice rather than a methodological advance \cite{recad}.

\begin{figure}[t]
\centering
\includegraphics[width=1.0\columnwidth]{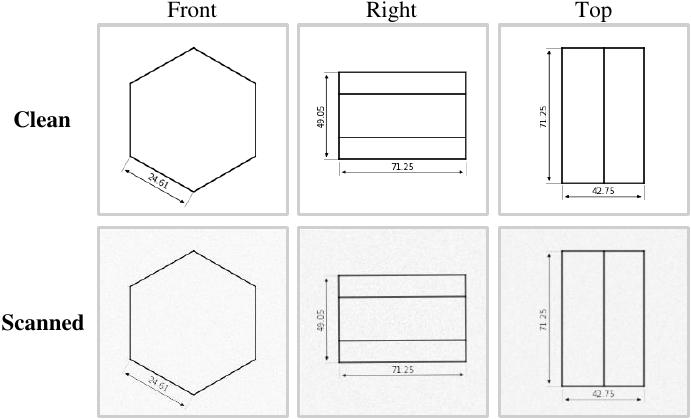}
\caption{Comparison between clean drawings and drawings transformed to resemble scanned drawings.}
\label{fig6}
\end{figure}

\subsection{Robustness to Scanned Drawings}
Industrial settings contain many drawings that have undergone scanning or printing, including ink bleeding, paper texture, and artifacts. To evaluate robustness under such conditions, we measure performance by providing drawings transformed to resemble scanned drawings as input, without any retraining. Specifically, we transform the Front, Right, and Top drawings by adding ink bleeding, paper texture noise, uneven brightness, artifacts, and rotation. A comparison between clean drawings without noise and the transformed drawings is shown in Figure~\ref{fig6}.

\begin{table}[t]
  \centering
  \setlength{\tabcolsep}{1.0mm}
  \small
  \begin{tabular}{ccccc}
    \toprule
    Method & ACC$_{cmd} \uparrow$ & ACC$_{param} \uparrow$ & MCD $\downarrow$ & IR $\downarrow$ \\
    \midrule
    CAD2Program    & 83.66 & 77.71 & 12.04 & 2.34 \\
    Vanilla    & 65.76 & 47.60 & 17.27 &  10.42\\
    + Cross-Attn & 85.33 & 80.07 & 8.48 & 1.43 \\
    + Sink & 85.46 & 80.19 & 8.37 & 1.30 \\
    \textbf{Ours (+AGL)}   & \textbf{86.32} & \textbf{81.02} & \textbf{8.15} & \textbf{0.99} \\
    \bottomrule
  \end{tabular}
  \caption{Results on drawings transformed to resemble scanned drawings, evaluated without retraining.}
  \label{tab4}
\end{table}

Table~\ref{tab4} shows the quantitative results under these conditions. From + Cross-Attn to Ours (+AGL), the degradation is small across all metrics compared to the results on clean drawings in Table~\ref{tab2}. Vanilla, in contrast, degrades substantially. Since Vanilla processes the geometry layer and the annotation layer together with a single image encoder, it appears unable to properly recognize either type of information when noise is present in the drawing. Drawing-Recode, on the other hand, extracts the annotation layer separately through YOLO and SVTR, which remain accurate under such noise without any retraining (98.18\% detection F1, 95.14\% word accuracy).

CAD2Program holds up far better than Vanilla in both ACC metrics, as its vision-language backbone is pretrained on large-scale OCR data spanning diverse image conditions \cite{internvl1.5}. However, relative to the clean drawings in Table~\ref{tab1}, its MCD rises by 18.9\%, against only 3.7\% for Drawing-Recode. In other words, CAD2Program reads the dimensions well owing to its pretrained OCR ability, but it appears difficult to fully capture both the annotation layer and the geometry layer with a single encoder. Drawing-Recode instead grounds the separately extracted annotation layer to the geometry layer through cross-attention and the AGL, keeping the correspondence between dimensions and geometry relatively stable even under noisy conditions. This suggests that our design is more robust to industrial drawing conditions than a single-encoder design.

\section{Conclusion}
We proposed Drawing-Recode, a framework that generates Parametric CAD code from raster-format 2D CAD drawings. By independently extracting the geometry layer and the annotation layer and explicitly grounding the two layers through the Annotation Grounding Loss, Drawing-Recode simultaneously addresses the dependence on vector formats and the insufficient learning of the correspondence between the two layers, both of which limit existing methods. Experiments show that Drawing-Recode outperforms the baselines across all metrics. It also maintains stable performance under conditions resembling scanned drawings, confirming that an architecture that processes the geometry layer and the annotation layer separately is robust to industrial drawing conditions.
\subsubsection{Limitation.}
Our work is based on the DeepCAD dataset, which covers diverse mechanical parts but is limited in modeling operations, and may therefore not fully reflect the more complex parts found in real industrial settings. In addition, our experiments on scanned drawings were conducted under artificially constructed conditions, and we did not directly handle scanned drawings collected from actual industrial settings. In future work, we plan to extend our approach using datasets containing more diverse and complex industrial CAD parts, as well as real scanned drawings.

\bibliography{aaai2027}

@misc{brepcoder,
  author        = {Kim, Mingi and Kim, Yongjun and Kang, Jungwoo and Kim, Hyungki},
  title         = {BrepCoder: A Unified Multimodal Large Language Model for Multi-task {B}-rep Reasoning},
  year          = {2026},
  eprint        = {2602.22284},
  archivePrefix = {arXiv}
}

@inproceedings{recad,
  author    = {Li, J. and Luo, Y. and Lou, Y. and Zhou, X.},
  title     = {{ReCAD}: Reinforcement Learning Enhanced Parametric {CAD} Model Generation with Vision-Language Models},
  booktitle = {Proceedings of the AAAI Conference on Artificial Intelligence},
  volume    = {40},
  number    = {8},
  pages     = {6190--6198},
  year      = {2026}
}

@inproceedings{cad-recode,
  author    = {Rukhovich, D. and Dupont, E. and Mallis, D. and Cherenkova, K. and Kacem, A. and Aouada, D.},
  title     = {{CAD-Recode}: Reverse Engineering {CAD} Code from Point Clouds},
  booktitle = {Proceedings of the IEEE/CVF International Conference on Computer Vision (ICCV)},
  pages     = {9801--9811},
  year      = {2025}
}

@misc{cadrille,
  author        = {Kolodiazhnyi, M. and Tarasov, D. and Zhemchuzhnikov, D. and Nikulin, A. and Zisman, I. and Vorontsova, A. and others},
  title         = {{cadrille}: Multi-Modal {CAD} Reconstruction with Online Reinforcement Learning},
  year          = {2025},
  eprint        = {2505.22914},
  archivePrefix = {arXiv}
}

@misc{text-to-cadquery,
  author        = {Xie, H. and Ju, F.},
  title         = {{Text-to-CadQuery}: A New Paradigm for {CAD} Generation with Scalable Large Model Capabilities},
  year          = {2025},
  eprint        = {2505.06507},
  archivePrefix = {arXiv}
}

@inproceedings{cad-coder,
  author    = {Guan, Y. and Wang, X. and Xing, X. and Zhang, J. and Xu, D. and Yu, Q.},
  title     = {{CAD-Coder}: Text-to-{CAD} Generation with Chain-of-Thought and Geometric Reward},
  booktitle = {Advances in Neural Information Processing Systems},
  volume    = {38},
  pages     = {59765--59789},
  year      = {2025}
}

@inproceedings{cad-llama,
  author    = {Li, J. and Ma, W. and Li, X. and Lou, Y. and Zhou, G. and Zhou, X.},
  title     = {{CAD-Llama}: Leveraging Large Language Models for Computer-Aided Design Parametric {3D} Model Generation},
  booktitle = {Proceedings of the IEEE/CVF Conference on Computer Vision and Pattern Recognition (CVPR)},
  pages     = {18563--18573},
  year      = {2025}
}

@misc{llama,
  author        = {Touvron, H. and Lavril, T. and Izacard, G. and Martinet, X. and Lachaux, M. A. and Lacroix, T. and others},
  title         = {{LLaMA}: Open and Efficient Foundation Language Models},
  year          = {2023},
  eprint        = {2302.13971},
  archivePrefix = {arXiv}
}

@article{algorithm1,
  author  = {Wang, W. and Grinstein, G. G.},
  title   = {A Survey of {3D} Solid Reconstruction from {2D} Projection Line Drawings},
  journal = {Computer Graphics Forum},
  volume  = {12},
  number  = {2},
  pages   = {137--158},
  year    = {1993}
}

@article{algorithm2,
  author  = {Kuo, M. H.},
  title   = {Reconstruction of Quadric Surface Solids from Three-View Engineering Drawings},
  journal = {{Computer-Aided Design}},
  volume  = {30},
  number  = {7},
  pages   = {517--527},
  year    = {1998}
}

@article{algorithm3,
  author  = {Liu, S. X. and Hu, S. M. and Chen, Y. J. and Sun, J. G.},
  title   = {Reconstruction of Curved Solids from Engineering Drawings},
  journal = {{Computer-Aided Design}},
  volume  = {33},
  number  = {14},
  pages   = {1059--1072},
  year    = {2001}
}

@article{algorithm4,
  author  = {Gong, J. H. and Zhang, G. F. and Zhang, H. and Sun, J. G.},
  title   = {Reconstruction of {3D} Curvilinear Wire-Frame from Three Orthographic Views},
  journal = {Computers \& Graphics},
  volume  = {30},
  number  = {2},
  pages   = {213--224},
  year    = {2006}
}

@misc{photo2cad,
  author        = {Harish, A. B. and Prasad, A. R.},
  title         = {{Photo2CAD}: Automated {3D} Solid Reconstruction from {2D} Drawings Using {OpenCV}},
  year          = {2021},
  eprint        = {2101.04248},
  archivePrefix = {arXiv}
}

@article{free2cad,
  author  = {Li, C. and Pan, H. and Bousseau, A. and Mitra, N. J.},
  title   = {{Free2CAD}: Parsing Freehand Drawings into {CAD} Commands},
  journal = {ACM Transactions on Graphics},
  volume  = {41},
  number  = {4},
  pages   = {1--16},
  year    = {2022}
}

@inproceedings{plankassembly,
  author    = {Hu, W. and Zheng, J. and Zhang, Z. and Yuan, X. and Yin, J. and Zhou, Z.},
  title     = {{PlankAssembly}: Robust {3D} Reconstruction from Three Orthographic Views with Learnt Shape Programs},
  booktitle = {Proceedings of the IEEE/CVF International Conference on Computer Vision (ICCV)},
  pages     = {18495--18505},
  year      = {2023}
}

@inproceedings{drawing2cad,
  author    = {Qin, F. and Lu, S. and Hou, J. and Wang, C. and Fang, M. and Liu, L.},
  title     = {{Drawing2CAD}: Sequence-to-Sequence Learning for {CAD} Generation from Vector Drawings},
  booktitle = {Proceedings of the 33rd ACM International Conference on Multimedia},
  pages     = {10573--10582},
  year      = {2025}
}

@inproceedings{cad2program,
  author    = {Wang, X. and Zheng, J. and Hu, Y. and Zhu, H. and Yu, Q. and Zhou, Z.},
  title     = {From {2D} {CAD} Drawings to {3D} Parametric Models: A Vision-Language Approach},
  booktitle = {Proceedings of the AAAI Conference on Artificial Intelligence},
  volume    = {39},
  number    = {8},
  pages     = {7961--7969},
  year      = {2025}
}

@inproceedings{vit,
  author    = {Dosovitskiy, A. and Beyer, L. and Kolesnikov, A. and Weissenborn, D. and Zhai, X. and Unterthiner, T. and others},
  title     = {An Image is Worth 16x16 Words: Transformers for Image Recognition at Scale},
  booktitle = {International Conference on Learning Representations (ICLR)},
  year      = {2021}
}

@article{cad1,
  author  = {Zhou, J. and Camba, J. D.},
  title   = {The Status, Evolution, and Future Challenges of Multimodal Large Language Models ({LLMs}) in Parametric {CAD}},
  journal = {Expert Systems with Applications},
  volume  = {282},
  pages   = {127520},
  year    = {2025}
}

@article{cad2,
  author  = {Zou, Q. and Wu, Y. and Liu, Z. and Xu, W. and Gao, S.},
  title   = {Intelligent {CAD 2.0}},
  journal = {Visual Informatics},
  volume  = {8},
  number  = {4},
  pages   = {1--12},
  year    = {2024}
}

@inproceedings{deepcad,
  author    = {Wu, R. and Xiao, C. and Zheng, C.},
  title     = {{DeepCAD}: A Deep Generative Network for Computer-Aided Design Models},
  booktitle = {Proceedings of the IEEE/CVF International Conference on Computer Vision (ICCV)},
  pages     = {6772--6782},
  year      = {2021}
}

@article{fusion360,
  author  = {Willis, K. D. and Pu, Y. and Luo, J. and Chu, H. and Du, T. and Lambourne, J. G. and Matusik, W.},
  title   = {{Fusion 360} Gallery: A Dataset and Environment for Programmatic {CAD} Construction from Human Design Sequences},
  journal = {ACM Transactions on Graphics},
  volume  = {40},
  number  = {4},
  pages   = {1--24},
  year    = {2021}
}

@inproceedings{cad-parser,
  author    = {Zhou, S. and Tang, T. and Zhou, B.},
  title     = {{CADParser}: A Learning Approach of Sequence Modeling for {B-Rep} {CAD}},
  booktitle = {Proceedings of the International Joint Conference on Artificial Intelligence (IJCAI)},
  pages     = {1804--1812},
  year      = {2023}
}

@inproceedings{yolo,
  author    = {Redmon, J. and Divvala, S. and Girshick, R. and Farhadi, A.},
  title     = {{You Only Look Once}: Unified, Real-Time Object Detection},
  booktitle = {Proceedings of the IEEE Conference on Computer Vision and Pattern Recognition (CVPR)},
  pages     = {779--788},
  year      = {2016}
}

@misc{svtr,
  author        = {Du, Y. and Chen, Z. and Jia, C. and Yin, X. and Zheng, T. and Li, C. and Jiang, Y. G.},
  title         = {{SVTR}: Scene Text Recognition with a Single Visual Model},
  year          = {2022},
  eprint        = {2205.00159},
  archivePrefix = {arXiv}
}

@inproceedings{bert,
  author    = {Devlin, J. and Chang, M. W. and Lee, K. and Toutanova, K.},
  title     = {{BERT}: Pre-Training of Deep Bidirectional Transformers for Language Understanding},
  booktitle = {Proceedings of the Conference of the North American Chapter of the Association for Computational Linguistics: Human Language Technologies (NAACL-HLT)},
  pages     = {4171--4186},
  year      = {2019}
}

@inproceedings{clip,
  author    = {Radford, A. and Kim, J. W. and Hallacy, C. and Ramesh, A. and Goh, G. and Agarwal, S. and Sutskever, I.},
  title     = {Learning Transferable Visual Models from Natural Language Supervision},
  booktitle = {Proceedings of the International Conference on Machine Learning (ICML)},
  pages     = {8748--8763},
  year      = {2021}
}

@misc{qwen,
  author        = {Bai, J. and Bai, S. and Chu, Y. and Cui, Z. and Dang, K. and Deng, X. and Zhu, T.},
  title         = {{Qwen} Technical Report},
  year          = {2023},
  eprint        = {2309.16609},
  archivePrefix = {arXiv}
}

@misc{qwen2,
  author        = {Yang, An and Yang, Baosong and Hui, Binyuan and others},
  title         = {{Qwen2} Technical Report},
  year          = {2024},
  eprint        = {2407.10671},
  archivePrefix = {arXiv}
}

@misc{qwen2.5,
  author        = {Yang, An and Yang, Baosong and Zhang, Beichen and others},
  title         = {{Qwen2.5} Technical Report},
  year          = {2024},
  eprint        = {2412.15115},
  archivePrefix = {arXiv},
  primaryClass  = {cs.CL}
}

@misc{llama3,
  author        = {Grattafiori, A. and Dubey, A. and Jauhri, A. and Pandey, A. and Kadian, A. and Al-Dahle, A. and Vasic, P. and others},
  title         = {The {Llama 3} Herd of Models},
  year          = {2024},
  eprint        = {2407.21783},
  archivePrefix = {arXiv}
}

@article{cadcl,
  author  = {Liang, J. and He, F. and Fan, R. and Chu, Y. and Yan, X.},
  title   = {{CADCL}: Reconstruct Parametric {CAD} Models from {B-rep} via Contrastive Learning},
  journal = {Journal of Computational Design and Engineering},
  volume  = {12},
  number  = {10},
  pages   = {176--184},
  year    = {2025}
}

@inproceedings{svtrv2,
  author    = {Du, Y. and Chen, Z. and Xie, H. and Jia, C. and Jiang, Y. G.},
  title     = {{SVTRv2}: {CTC} Beats Encoder-Decoder Models in Scene Text Recognition},
  booktitle = {Proceedings of the IEEE/CVF International Conference on Computer Vision (ICCV)},
  pages     = {20147--20156},
  year      = {2025}
}

@inproceedings{cadcodeverify,
  author    = {Alrashedy, K. and Tambwekar, P. and Zaidi, Z. H. and Langwasser, M. and Xu, W. and Gombolay, M.},
  title     = {Generating {CAD} Code with Vision-Language Models for {3D} Designs},
  booktitle = {Proceedings of the International Conference on Learning Representations (ICLR)},
  pages     = {52236--52262},
  year      = {2025}
}

@inproceedings{yolov10,
  author    = {Wang, A. and Chen, H. and Liu, L. and Chen, K. and Lin, Z. and Han, J. and Ding, G.},
  title     = {{YOLOv10}: Real-Time End-to-End Object Detection},
  booktitle = {Advances in Neural Information Processing Systems (NeurIPS)},
  volume    = {37},
  pages     = {107984--108011},
  year      = {2024}
}

@article{cad-survey,
  author  = {Zhang, L. and Le, B. and Akhtar, N. and Lam, S. K. and Ngo, D.},
  title   = {Large Language Models for Computer-Aided Design: A Survey},
  journal = {ACM Computing Surveys},
  volume  = {58},
  number  = {9},
  pages   = {1--39},
  year    = {2026}
}

@misc{futurecad,
  author        = {Li, J. and Zhang, Q. and Chen, Q. and Qiu, G. and Lou, Y. and Zhou, X.},
  title         = {Towards High-Fidelity {CAD} Generation via {LLM}-Driven Program Generation and Text-Based {B-Rep} Primitive Grounding},
  year          = {2026},
  eprint        = {2603.11831},
  archivePrefix = {arXiv},
  primaryClass  = {cs.CV}
}

@inproceedings{internvl,
  author    = {Chen, Z. and Wu, J. and Wang, W. and Su, W. and Chen, G. and Xing, S. and others},
  title     = {{InternVL}: Scaling Up Vision Foundation Models and Aligning for Generic Visual-Linguistic Tasks},
  booktitle = {Proceedings of the IEEE/CVF Conference on Computer Vision and Pattern Recognition (CVPR)},
  pages     = {24185--24198},
  year      = {2024}
}

@article{mini-internvl,
  author  = {Gao, Z. and Chen, Z. and Cui, E. and Ren, Y. and Wang, W. and Zhu, J. and others},
  title   = {{Mini-InternVL}: A Flexible-Transfer Pocket Multi-Modal Model with 5\% Parameters and 90\% Performance},
  journal = {Visual Intelligence},
  volume  = {2},
  number  = {1},
  pages   = {32},
  year    = {2024}
}

@article{internvl1.5,
  author  = {Chen, Z. and Wang, W. and Tian, H. and Ye, S. and Gao, Z. and Cui, E. and others},
  title   = {How Far Are We to {GPT-4V}? Closing the Gap to Commercial Multimodal Models with Open-Source Suites},
  journal = {Science China Information Sciences},
  volume  = {67},
  number  = {12},
  pages   = {220101},
  year    = {2024}
}


\end{document}